\documentclass[letterpaper, 10 pt, conference]{ieeeconf}

\IEEEoverridecommandlockouts
\usepackage{cite}
\usepackage{amsmath,amssymb,amsfonts}
\usepackage{algorithmic}
\usepackage{graphicx}
\usepackage{textcomp}
\usepackage[table]{xcolor}
\usepackage{comment}
\usepackage[ruled,vlined]{algorithm2e}
\usepackage[font=small]{caption}
\usepackage{array}
\usepackage{subcaption}
\usepackage{arydshln}

\usepackage{multirow}

\usepackage{tensor}
\newcommand{\eg}{\emph{e.g.},}
\newcommand{\ie}{\emph{i.e.},}

\usepackage{float}
\usepackage{rotating}
\usepackage{nicematrix}  %
\usepackage{mathtools}
\usepackage{amsmath}
\usepackage{pifont}
\usepackage{cuted}
\usepackage{dblfloatfix}
\usepackage{balance}
\usepackage{titlesec}

\let\labelindent\relax
\usepackage{enumitem}

\usepackage[pagebackref=true,breaklinks=true,letterpaper=true,colorlinks,bookmarks=false]{hyperref}

\usepackage{dsfont}
\usepackage{mathtools}
\usepackage{booktabs}
\usepackage{graphics}
\usepackage[mathscr]{euscript}

\newcommand{\cmark}{\ding{51}}%
\newcommand{\xmark}{\ding{55}}%

\newcommand{\algstep}[1]{\item[]\medskip\hrule\kern 2pt\hbox to \textwidth{\hspace{\labelsep}\textbf{#1}\hfill}\hrule}

\usepackage{xspace}
\makeatletter
\DeclareRobustCommand\bmvaOneDot{\futurelet\@let@token\bmv@onedotaux}
\def\bmv@onedotaux{\ifx\@let@token.\else.\null\fi\xspace}
\makeatother

\definecolor{LightCyan}{rgb}{0.88,1,0.88}
\definecolor{linear_color}{RGB}{220,223,240}
\definecolor{gray_bbox_color}{RGB}{243,243,244}
\newcommand{\coolname}{\text{CoHiD}}
\definecolor{rebuttal}{rgb}{0,0,1}

\def\eqref#1{Eq.~(\ref{#1})}

\newcommand\blfootnote[1]{%
  \begingroup
  \renewcommand\thefootnote{}\footnote{#1}%
  \addtocounter{footnote}{-1}%
  \endgroup
}

\begin{document}

\bstctlcite{IEEEexample:BSTcontrol}

\title{\LARGE \bf \coolname{}: Confidence-Aware Hyperbolic Image-to-LiDAR Distillation \\for LiDAR Representation Learning}

\author{Kaushik Roy$^{1}$, Saimunur Rahman$^{1}$, Sutharsan Mahendran$^{1,2}$, \\David Hall$^{1}$, Sagun Singh Shrestha$^{1}$, Peyman Moghadam$^{1,2}$
}

\maketitle

\begin{abstract}
Learning robust LiDAR representations is critical for autonomous systems, yet dense 3D annotation is expensive and difficult to scale. Image-to-LiDAR distillation offers a scalable alternative by transferring semantic knowledge from pre-trained Vision Foundation Models without requiring annotated 3D data. However, existing approaches typically align corresponding image and LiDAR features in Euclidean space and treat visual supervision uniformly, despite substantial variation in semantic specificity and correspondence reliability. We introduce \coolname{}, a confidence-aware image-to-LiDAR distillation framework that operates in hyperbolic space. Specifically, we embed cross-modal features in the unit Poincaré ball and use hyperbolic radial distance as a confidence proxy, emphasizing informative teacher signals while suppressing uncertain ones. We further perform distillation at three complementary scales. Pixel-to-point alignment transfers fine-grained visual information, superpixel-to-superpoint alignment promotes region-level semantic consistency and reduces sensitivity to individual correspondence errors, and global alignment preserves scene-level semantic context. Together, these components enable selective and structured transfer of semantic knowledge from pre-trained vision foundation models to LiDAR representations. Extensive experiments on five large-scale autonomous driving benchmark settings demonstrate improved downstream LiDAR semantic segmentation, together with stronger robustness under corruptions and cross-domain shifts.

\blfootnote{$^1$ CSIRO Robotics, CSIRO, Australia. 
E-mails: {\tt\footnotesize \emph{firstname.lastname}@csiro.au}}
\blfootnote{$^2$ Queensland University of Technology (QUT), Brisbane, Australia. 
E-mails: {\tt\footnotesize \emph{firstname.lastname}@qut.edu.au}}

\end{abstract}

\section{Introduction}

Robust LiDAR based perception is fundamental to modern robotic systems, enabling accurate 3D scene understanding for autonomous driving and embodied intelligence. In particular, 3D semantic segmentation plays a central role by providing dense semantic interpretations of complex environments, which is critical for safe navigation, obstacle avoidance, and high-level decision making~\cite{muhammad2020deep}. %
While LiDAR offers precise geometric measurements and robustness to illumination changes, learning semantically rich 3D representations typically requires dense point-wise annotations, which are costly and time-consuming to produce, making such supervision difficult to scale~\cite{gao2021we}. This scalability challenge has motivated growing interest in label-free representation learning for LiDAR perception, where supervision is obtained without relying on manually annotated 3D labels.

Self-supervised pre-training has emerged as a promising paradigm to address this bottleneck, allowing models to learn rich representations from abundant unlabeled data. One way to achieve such label-free supervision is through cross-modal knowledge distillation, where a stronger modality provides supervisory signals for another without requiring explicit annotations. In particular, knowledge distillation from 2D images to 3D LiDAR offers a scalable solution by leveraging the vast progress in Vision Foundation Models (VFMs), such as DINOv2~\cite{oquab2024dinov}, which provide strong semantic features pretrained on massive image datasets without labels~\cite{kang2025domain}. By projecting image features onto corresponding LiDAR points, these methods transfer rich 2D semantic knowledge to the 3D domain in a label-free manner and have demonstrated strong performance on downstream tasks such as semantic segmentation and object detection, thereby significantly reducing the need for large-scale annotated 3D data~\cite{ccanakcci2025label}.

Despite recent advances, image-to-LiDAR distillation remains limited by both the reliability of cross-modal supervision and the geometry used to transfer semantic knowledge. Existing methods typically align visual features with corresponding LiDAR points or regions based on sensor calibration~\cite{sautier2022image,puy2024three,mahendran2026tolid}. However, these correspondences are not uniformly reliable, as occlusions, viewpoint differences, object motion, and projection inaccuracies can introduce ambiguous or mismatched supervision. At the same time, conventional distillation objectives primarily encourage similarity between paired 2D and 3D features in standard feature spaces, without explicitly accounting for the relationships among semantic concepts. This is particularly limiting in outdoor scenes, where categories often exhibit coarse-to-fine relationships, such as vehicles comprising cars and trucks or vegetation comprising trees and bushes, and where the transferred visual evidence can vary considerably in semantic specificity. These observations motivate a hyperbolic formulation of image-to-LiDAR distillation that can better preserve the underlying semantic structure of visual representations while providing a more suitable representation for handling variations in the reliability and specificity of cross-modal supervision.

\begin{figure*}[t]
\centering
\includegraphics[width=1\textwidth, scale=1]{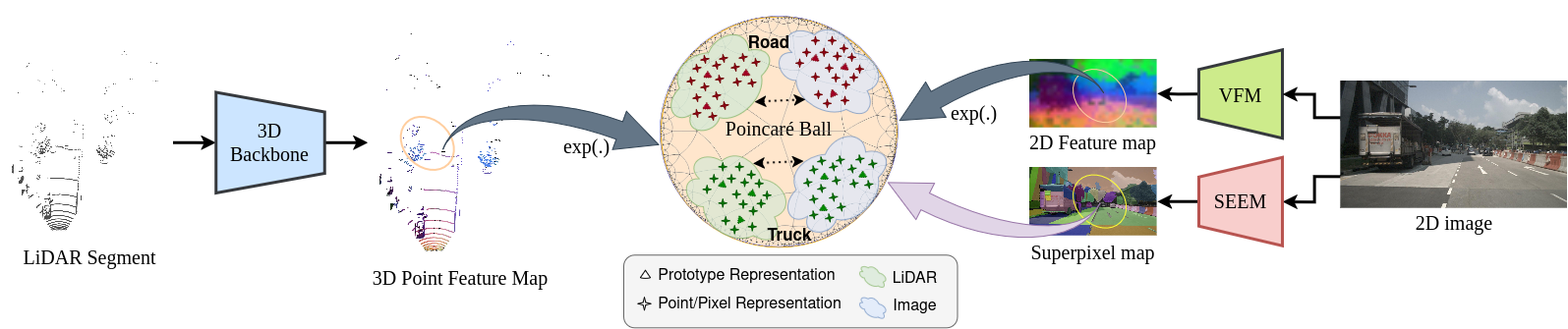}
\vspace{-1.0em}
\caption{Overview of \coolname{}, our confidence-aware hierarchical distillation framework. Euclidean features from a 2D vision foundation model and a 3D LiDAR network are projected into the Poincaré ball. The hyperbolic radial distance provides a geometry-aware confidence measure to gate the distillation loss. Alignment is performed simultaneously at the fine-grained (pixel-to-point) and region-level (superpixel-to-superpoint) scales, ensuring robust semantic consistency.}
\label{fig:method}
\vspace{-1.5em}
\end{figure*}

In this work, we propose \coolname{}, a confidence-aware image-to-LiDAR distillation framework that transfers visual knowledge to LiDAR representations in hyperbolic space. Specifically, we adopt the unit Poincaré ball, whose negatively curved geometry provides a natural representation for structured semantic relationships~\cite{fang2023poincare}. Beyond modeling semantic structure, hyperbolic embeddings exhibit a useful radial organization, where more specific and confident representations tend to lie farther from the origin, while generic or ambiguous representations remain closer to it. We exploit this property to estimate the reliability of visual supervision during distillation. Given a pre-trained VFM teacher (\eg{} DINOv2), we project the corresponding image and LiDAR features into the Poincaré ball and align them using hyperbolic distance, while weighting each correspondence according to the radial position of the teacher embedding. This allows the model to emphasize semantically informative visual features and suppress ambiguous supervision. To capture complementary levels of information, we further perform distillation at three scales. \textit{Pixel-to-point alignment} transfers fine-grained visual information, \textit{superpixel-to-superpoint alignment} promotes region-level semantic consistency and reduces sensitivity to individual correspondence errors, and \textit{global alignment} preserves scene-level semantic structure. Together, these components enable robust, label-free transfer of semantic knowledge from pre-trained VFMs to LiDAR representations for downstream 3D perception task. An overview of \coolname{} is shown in Figure~\ref{fig:method}.

Extensive experiments on five public large-scale autonomous driving benchmarks show that \coolname{} improves LiDAR semantic segmentation performance. It also enhances robustness under corruptions and cross-domain shifts, highlighting the effectiveness of confidence-aware hyperbolic distillation for LiDAR representation learning.
Our main contributions are summarized as follows: 

\begin{itemize}[leftmargin=*]
\item We introduce a confidence-aware formulation of image-to-LiDAR
distillation in hyperbolic space, where the radial position of visual embeddings provides a continuous confidence proxy for selectively modulating the contribution of teacher signals during cross-modal transfer.

\item We develop hierarchical cross-modal distillation approach at the pixel-to-point, superpixel-to-superpoint, and global levels, jointly capturing fine-grained correspondences, region-level semantic consistency, and scene-level context.

\item We evaluate \coolname{} on five large-scale benchmarks, demonstrating improved LiDAR representations and stronger robustness to corruptions and cross-domain shifts.
\end{itemize}

\section{Related Work}

Learning transferable LiDAR representations has become increasingly important for reducing the dependence of 3D perception systems on large-scale task-specific annotations. Conventional LiDAR perception methods typically learn representations under full supervision using point-based~\cite{Puy2023WI}, voxel-based~\cite{hong2021lidar}, range-view~\cite{ouyang2025mdc}, bird's-eye-view~\cite{zhang2020polarnet}, or multi-view architectures~\cite{zhuang2021perception}. Although these approaches provide effective representations for downstream task, obtaining dense point-level annotations remains expensive and time-consuming because of the sparsity and geometric complexity of LiDAR data~\cite{gao2021we}. This has motivated label-efficient representation learning and, more recently, the use of foundation models to generate scalable supervision for 3D data~\cite{ccanakcci2025label}.

Self-supervised learning provides a natural alternative by pretraining LiDAR encoders directly from unlabeled data. Early 3D representation learning methods have largely focused on object-centric~\cite{sanghi2020info3d} or indoor settings~\cite{chen20224dcontrast}, using geometric pretext tasks~\cite{zhang2025self}, contrastive learning~\cite{xie2020pointcontrast,nunes2022segcontrast}, or masked modeling~\cite{wu2023masked} to construct supervision without manual labels. For example, {PointContrast}~\cite{xie2020pointcontrast} learns point-level representations by maximizing agreement between geometrically transformed observations, while {SegContrast}~\cite{nunes2022segcontrast} introduces segmentation-oriented contrastive objectives to encourage locally consistent features. Temporal information has also been exploited to improve representation consistency across sequential observations~\cite{dewan2020deeptemporalseg}. Despite enabling label-free pretraining, purely 3D objectives derive their supervision predominantly from geometric relationships within the point cloud and therefore have limited access to the rich semantic knowledge encoded by large-scale VFMs.

Image-to-LiDAR distillation addresses this limitation by transferring semantic representations from pre-trained image encoders to LiDAR backbones using synchronized camera and LiDAR observations. {PPKT}~\cite{liu2021learning} introduced fine-grained pixel-to-point contrastive transfer, aligning individual image features with their projected 3D counterparts. {SLidR}~\cite{sautier2022image} instead performs superpixel-to-superpoint distillation, providing more spatially coherent supervision and reducing sensitivity to unreliable individual correspondences. Subsequent methods have introduced additional spatial and temporal constraints to strengthen cross-modal representation learning. FOr example, {Seal}~\cite{liu2023segment} exploits vision foundation models together with spatial and temporal regularization, while {SuperFlow}~\cite{xu20244d} incorporates dense temporal correspondence across consecutive observations. Recent work has further emphasized the importance of scale and representation diversity. {LargeAD}~\cite{kong2025largead} extends VFM-guided pretraining across heterogeneous autonomous-driving datasets, while {LiMoE}~\cite{xu2025limoe} combines complementary range-, voxel-, and point-based LiDAR representations through a mixture-of-experts framework. {ScaLR}~\cite{puy2024three} further demonstrates that stronger visual teachers, larger LiDAR backbones, and diverse pretraining data substantially improve the transferability and generalization of learned LiDAR representations.

Despite this progress, existing image-to-LiDAR representation learning has primarily advanced through improved cross-modal correspondences, spatial and temporal consistency, stronger visual teachers, and larger-scale pretraining. However, the reliability of individual teacher representations is rarely considered explicitly during knowledge transfer, even when their cross-modal correspondences may be affected by occlusion, dynamic objects, or projection ambiguity. Moreover, existing methods predominantly optimize feature similarity in conventional embedding spaces without explicitly modeling the structured semantic relationships encoded by VFMs. These limitations leave the reliability and geometry of cross-modal knowledge transfer comparatively underexplored, which is effectively addressed by \coolname{}.

\section{Methodology}

This section presents \coolname{} for learning robust LiDAR representations without manual 3D annotations. We first establish the mathematical foundations of hyperbolic geometry and the Poincaré ball model, which naturally encode semantic hierarchies. We then describe confidence-aware distillation from a frozen VFM to a LiDAR student at three complementary levels, namely pixel-to-point, superpixel-to-superpoint, and global scene alignment.

\subsection{Geometric Foundations}

We formulate cross-modal feature alignment in hyperbolic space to better preserve structured semantic relationships present in visual representations. A hyperbolic space is a Riemannian manifold with constant negative curvature. Its volume grows exponentially with distance from the origin, allowing representations with hierarchical or tree-like organization to be embedded with lower distortion than in Euclidean space. We adopt the Poincaré ball model due to its conformality and computational convenience.

Formally, a $d$-dimensional hyperbolic space with curvature $c$ (where $c>0$) is denoted as a Riemannian manifold $(\mathcal{M},g)$, where  $\mathcal{M}$ denotes the underlying differentiable manifold, and $g$ is the Riemannian metric that dictates distances and angles. For deep learning applications, the Poincaré ball model $\mathcal{B}_c^d$ is highly effective due to its conformality and computational tractability~\cite{nickel2017poincare}. The $d$-dimensional Poincaré ball of curvature $c$ is defined as:
\begin{equation}
\mathcal{B}_c^d=\Big\{\mathbf{x}\in\mathbb{R}^d\mid\|\mathbf{x}\|<\frac{1}{\sqrt{c}}\Big\},
\end{equation}
where $\|\cdot\|$ denotes the standard Euclidean norm. The distance between any two points $\mathbf{x},\mathbf{y}\in\mathcal{B}_c^d$ is computed via the hyperbolic metric:
\begin{equation}
d_{\mathcal{B}}(\mathbf{x},\mathbf{y})=\frac{2}{\sqrt{c}}\mathrm{arctanh}\Big(\sqrt{c}\|-\mathbf{x}\oplus_c\mathbf{y}\|\Big).
\end{equation}

Here, $\oplus_c$ represents Möbius addition, a critical operation that ensures closure within the ball, defined as:
\begin{equation}
\mathbf{x}\oplus_c\mathbf{y}=\frac{(1+2c\langle\mathbf{x},\mathbf{y}\rangle+c\|\mathbf{y}\|^2)\mathbf{x}+(1-c\|\mathbf{x}\|^2)\mathbf{y}}{1+2c\langle\mathbf{x},\mathbf{y}\rangle+c^2\|\mathbf{x}\|^2\|\mathbf{y}\|^2},
\end{equation}
where $\langle\cdot,\cdot\rangle$ is the Euclidean inner product. To maintain numerical stability and ensure all embeddings remain strictly valid within the manifold, we apply a clipping projection during training: %
\begin{equation}
\operatorname{proj}(\mathbf{x})\gets\begin{cases}\mathbf{x},&\|\mathbf{x}\|<\frac{1}{\sqrt{c}}-\epsilon\\\frac{\mathbf{x}}{\|\mathbf{x}\|}\left(\frac{1}{\sqrt{c}}-\epsilon\right),&\text{otherwise},\end{cases}
\end{equation}
with $\epsilon>0$ acting as a small numerical boundary margin.

\begin{figure}[t]
\centering
\includegraphics[width=.38\textwidth, scale=1]{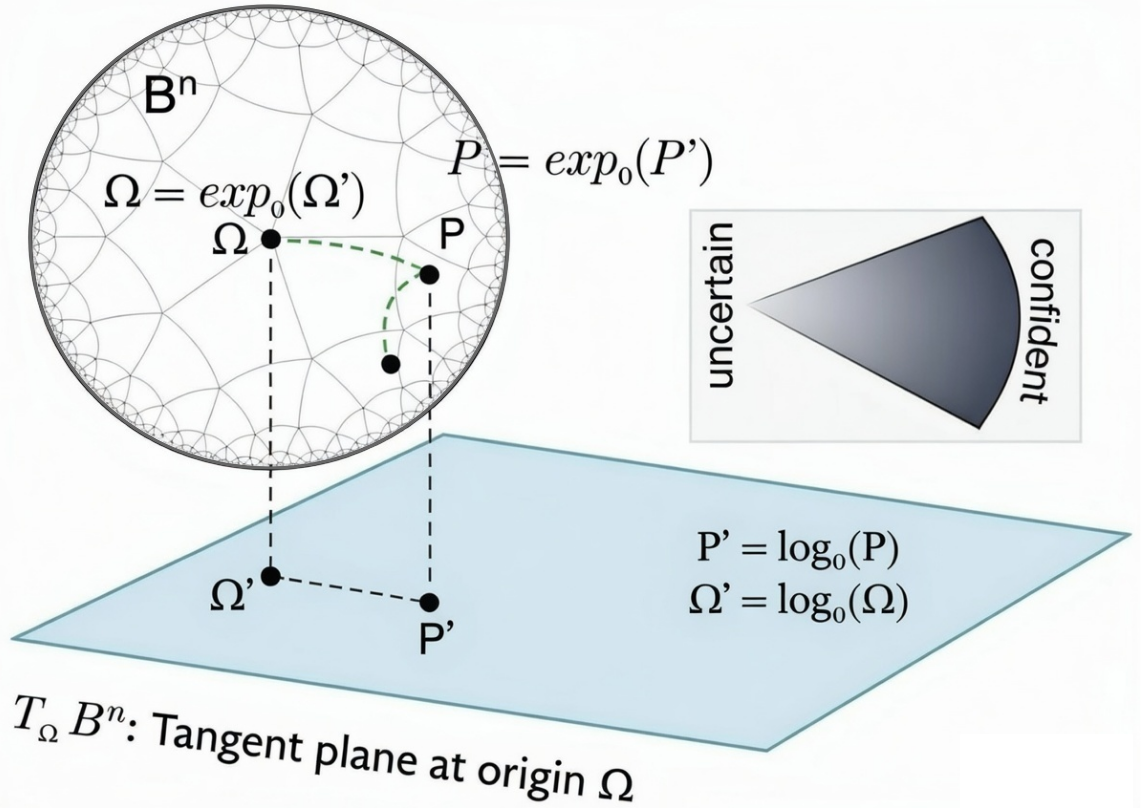}
\caption{Mapping between the Poincaré ball and the tangent space at the origin: points $\Omega$ and $P$ are mapped to the tangent space via $\log_0(\cdot)$ and mapped back using $\exp_0(\cdot)$. The hyperbolic radius serves as a confidence measure that increases toward the boundary of the ball.}
\label{fig:ball}
\vspace{-1.5em}
\end{figure}

Since direct feature aggregation in hyperbolic space is non-trivial as the Fréchet mean requires iterative optimization with no closed-form solution, we propose to project points to the tangent space at the origin $\mathbf{0}\in\mathcal{B}_c^d$, denoted $T_\mathbf{0}\mathcal{B}_c^d\cong\mathbb{R}^d$. This mapping is achieved using the logarithmic map (see Fig.~\ref{fig:ball}):
\begin{equation}
\log_\mathbf{0}^c(\mathbf{x})=\frac{2}{\sqrt{c}}\mathrm{arctanh}\big(\sqrt{c}\|\mathbf{x}\|\big)\frac{\mathbf{x}}{\|\mathbf{x}\|},\quad\mathbf{x}\neq\mathbf{0}.
\end{equation}

Once standard Euclidean operations (such as averaging) are performed in the tangent space, the resulting vectors are mapped back to the Poincaré ball using the exponential mapping:
\begin{equation}
\exp_\mathbf{0}^c(\mathbf{v})=\tanh\Big(\frac{\sqrt{c}\|\mathbf{v}\|}{2}\Big)\frac{\mathbf{v}}{\sqrt{c}\|\mathbf{v}\|},\quad\mathbf{v}\neq\mathbf{0}.
\end{equation}

\subsection{Confidence-Aware Hierarchical Distillation}

Now, we introduce our \underline{co}nfidence-aware \underline{hi}erarchical \underline{d}istillation (\coolname{}) framework that explicitly accounts for visual noise and semantic structure. %

We extract pixel-wise image features $\mathbf{f}^I\in\mathbb{R}^{N_I\times C_I}$ from a frozen 2D VFM (\ie{} DINOv2 \cite{oquab2024dinov}) acting as our teacher network, and point-wise LiDAR features $\mathbf{f}^L\in\mathbb{R}^{N_L\times C_L}$ from a 3D LiDAR backbone (\ie{} MinkUnet-34 \cite{choy20194d}) serving as the student. To enable cross-modal distillation, we project the LiDAR features into the same embedding dimension as the image features using a lightweight MLP projector, yielding $\tilde{\mathbf{f}}^L \in \mathbb{R}^{N_L \times C_I}$. Pixel-to-point correspondences are established via sensor calibration. 

We embed the Euclidean features into the unit Poincaré ball $\mathcal{B}_c^d$ by preserving their direction while mapping their unbounded norms to a bounded radius. Specifically, we use
\begin{equation} 
r(\mathbf{f})=R_{\max}\cdot\frac{|\mathbf{f}|_2}{|\mathbf{f}|_2+\tau}, 
\end{equation}
corresponding to $\tau=1$ and $R_{\max}=1$, followed by the exponential map at the origin, $\mathbf{z}=\exp_0^c\left(r(\mathbf{f})\frac{\mathbf{f}}{|\mathbf{f}|_2}\right)$.

In the Poincaré ball, the radial distance from the origin naturally encodes semantic specificity. Points near the boundary represent highly discriminative, confident features, while points near the origin correspond to ambiguous or noisy signals. We mathematically define the confidence of a hyperbolic embedding $\mathbf{z}$ based on its radial distance:
\begin{equation}
u(\mathbf{z})=\frac{1}{\sqrt{c}}\operatorname{arctanh}\left(\sqrt{c}\|\mathbf{z}\|_2\right),\quad\|\mathbf{z}\|_2<1.
\end{equation}
This confidence metric is then mapped to a %
radial gating weight, $w(\mathbf{z})=\min\big(1,u(\mathbf{z})\big)$.
During distillation, these weights are detached from the gradient. This ensures that reliable, structurally significant features dominate the learning process, effectively suppressing the impact of occlusions or motion blur inherited from the visual domain.

To enforce both local geometric alignment and global semantic consistency, we structure our distillation loss across three scales. At the fine-grained level, we apply a \textit{pixel-to-point} distillation loss on the last two feature layers, aligning corresponding elements while modulating their contribution by the teacher's confidence:
\begin{equation}
\mathcal{L}_{pp}=\mathbb{E}_{i}\left[w(\mathbf{z}^{I,l}_i)\,d_{\mathbb{B}}(\mathbf{z}^{I,l}_i,\mathbf{z}^{L,l}_i)\right].
\end{equation}
where $\mathbf{z}^{I,l}_i$ and $\mathbf{z}^{L,l}_i$ denote the image and LiDAR embeddings of the $i$-th corresponding pair at layer $l$, respectively.

At the coarser level, we group both modalities into shared superpixel and superpoint regions, yielding region assignments \(\mathcal{S}=\{s_i\}_{i=1}^N\), where each \(s_i\) denotes the region index of the \(i\)-th element. Let \(k\) index these regions and let \(\mathcal{S}_k=\{i:s_i=k\}\) be the set of elements belonging to region \(k\). To obtain a region-level prototype without distorting the manifold geometry, we first average the corresponding features in the tangent space at the origin and then map the result back to the hyperbolic space:
\begin{equation}
\bar{\mathbf{v}}_k=\frac{1}{|\mathcal{S}_k|}\sum_{i\in\mathcal{S}_k}\log_0(\mathbf{z}_i),\quad\bar{\mathbf{z}}_k=\exp_0(\bar{\mathbf{v}}_k).
\end{equation}
Here \(\bar{\mathbf{v}}_k\) is the Euclidean mean of the logarithmically mapped features of region \(k\), and \(\bar{\mathbf{z}}_k\) is the resulting hyperbolic prototype. The region-level \textit{superpixel-to-superpoint} loss is then defined as
\begin{equation}
\mathcal{L}_{ss}=\mathbb{E}_{k}\left[w(\bar{\mathbf{z}}^I_k)\,d_{\mathbb{B}}(\bar{\mathbf{z}}^I_k,\bar{\mathbf{z}}^L_k)\right].
\end{equation}

Beyond local point--pixel correspondences and region-level prototypes, we further distill scene-level semantic context through global feature summaries. Specifically, let $\mathbf{g}^{L}$ and $\mathbf{g}^{I}$ denote the max-pooled projected LiDAR feature and the DINO CLS token, respectively. We formulate the \textit{global} distillation objective in the Poincaré ball $\mathbb{B}^d_c$ as

\begin{equation}
\mathcal{L}_{ctx} = 
\mathbb{E}
\left[
w(\mathbf{g}^{I})d_{\mathbb{B}}(\mathbf{g}^{I},\mathbf{g}^{L})
\right],
\end{equation}
where $w(\mathbf{g}^{I})=\min\big(1,u(\mathbf{g}^{I})\big)$ is the radial gating weight determined by the teacher confidence metric $u(\mathbf{g}^I)$. This global objective complements the local and region-level losses by providing scene-level semantic supervision.

Our final training objective is the sum of these three components:
\begin{equation}
\mathcal{L}=\lambda_{pp} \mathcal{L}_{pp}+\lambda_{ss}\mathcal{L}_{ss}+\lambda_{ctx}\mathcal{L}_{ctx}.
\end{equation}

This hierarchical formulation combines fine-grained correspondence, region-level structure, and global scene semantics within a unified hyperbolic space. Tangent-space aggregation reduces the influence of localized visual noise when constructing region prototypes, while the hyperbolic geometry provides greater separation for semantically distinct representations, particularly near the boundary of the Poincaré ball. Together with confidence-aware weighting, these properties enable robust, label-free LiDAR representation learning that jointly captures semantic structure and teacher reliability.

\section{Experimental Setup}

\begin{table*}[t!]
\centering
\caption{\textbf{Performance} comparison (mIoU, \%) of state-of-the-art LiDAR pretraining methods pretrained on nuScenes and fine-tuned under varying label budgets on nuScenes, and SemanticKITTI. \textbf{Best} results are highlighted in \textbf{bold}. LP indicates linear probing with frozen feature extractors. All methods employ the MinkUNet-34~\cite{choy20194d} backbone and are pretrained solely on nuScenes.}
\label{tab:finetuning_benchmark}
\setlength{\tabcolsep}{9pt}
\small
\begin{tabular}{ccccccccccc}
\hline
\multirow{2}{*}{} &
\multirow{2}{*}{Method} & 
\multirow{2}{*}{Venue} & 
\multirow{2}{*}{Epochs} &
\multicolumn{6}{c}{nuScenes} &
\multicolumn{1}{c}{S.KITTI} \\
\cmidrule(lr){5-10} \cmidrule(lr){11-11}
& & & & LP & 1\% & 5\% & 10\% & 25\% & Full & 1\% \\
\hline
& Random  & - & - & 8.10 & 30.30 & 47.84 & 56.15 & 65.48 & 74.66 & 39.50   \\ 
 \cdashline{1-11} \noalign{\vskip 2pt}
\multirow{9}{*}{\rotatebox{90}{\textbf{ViT-B}}}
&
PPKT~\cite{liu2021learning}    & arXiv'21 & 50    
& 39.95 & 40.91 & 53.21 & 60.87 & 66.22 & 74.07 & 44.09 \\

&
SLidR~\cite{sautier2022image}    & CVPR'22 & 50 
& 45.35 & 41.64 & 55.83 & 62.68 & 67.61 & 74.98 & 45.50 \\

&
Seal~\cite{liu2023segment}    & NIPS'23 & 50 
& 46.59 & 45.98 & 57.15 & 62.79 & 68.18 & 75.41 & 47.24 \\

&
Superflow~\cite{xu20244d}    & ECCV'24 &  150
& 47.66 & 48.09 & 59.66 & 64.52 & 69.79 & 76.57 & 48.40 \\

&
ScaLR~\cite{puy2024three}    & CVPR'24 &  50
& 51.90 & 48.90 & 57.69 & 62.88 & 66.85 & 74.15 & 47.77 \\

&
LargeAD~\cite{kong2025largead}    & TPAMI'25 &  50
& 47.84 & 48.37 & 59.36 & 64.11 & 69.02 & 75.91 & 50.68 \\

&
Superflow++~\cite{xu2025enhanced}    & TPAMI'25 &  50
& 48.86 & 49.56 & 60.75 & 65.46 & 70.19 & 77.29 & 49.90 \\

&
LiMoE~\cite{xu2025limoe}    & CVPR'25 & 150 
& 49.07 & 50.23 & 61.51 & 66.17 & 71.56 & 77.81 & 50.30 \\

&
\cellcolor{green!15}\textbf{\coolname{}}~(Ours)   & \cellcolor{green!15} - 
& \cellcolor{green!15}\textbf{25}
& \cellcolor{green!15}\textbf{60.95}
& \cellcolor{green!15}\textbf{52.46}
& \cellcolor{green!15}\textbf{66.99}
& \cellcolor{green!15}\textbf{71.83}
& \cellcolor{green!15}\textbf{73.90}
& \cellcolor{green!15}\textbf{78.52}
& \cellcolor{green!15}\textbf{53.72} \\

\hline

\multirow{9}{*}{\rotatebox{90}{\textbf{ViT-L}}}
&
PPKT~\cite{liu2021learning}    & arXiv'21 &  50
& 41.57 & 42.05 & 55.75 & 61.26 & 66.88 & 74.33 & 45.87 \\

&
SLidR~\cite{sautier2022image}    & CVPR'22 &  50
& 45.70 & 42.77 & 57.45 & 63.20 & 68.13 & 75.51 & 47.01 \\

&
Seal~\cite{liu2023segment}    & NIPS'23 &  50
& 46.81 & 46.27 & 58.14 & 63.27 & 68.67 & 75.66 & 47.55 \\

&
Superflow~\cite{xu20244d}    & ECCV'24 &  150
& 48.01 & 49.95 & 60.72 & 65.09 & 70.01 & 77.19 & 49.07 \\

&
ScaLR~\cite{puy2024three}    & CVPR'24 &  50
& 51.77 & 49.13 & 58.36 & 62.75 & 66.80 & 74.16 & 48.64 \\

&
LargeAD~\cite{kong2025largead}    & TPAMI'25 &  50
& 48.71 & 49.21 & 60.37 & 64.82 & 69.85 & 75.94 & 51.68 \\

&
Superflow++~\cite{xu2025enhanced}    & TPAMI'25 &  50
& 49.78 & 50.92 & 61.83 & 66.30 & 71.07 & 77.63 & 50.33 \\

&
LiMoE~\cite{xu2025limoe}    & CVPR'25 &  150
& 49.35 & 51.41 & 62.07 & 66.64 & 71.59 & 77.85 & 50.69 \\

&
\cellcolor{green!15}\textbf{\coolname{}}~(Ours)     & \cellcolor{green!15} -
& \cellcolor{green!15}\textbf{25} 
& \cellcolor{green!15}\textbf{60.07}
& \cellcolor{green!15}\textbf{53.29}
& \cellcolor{green!15}\textbf{65.74}
& \cellcolor{green!15}\textbf{70.30}
& \cellcolor{green!15}\textbf{73.53}
& \cellcolor{green!15}\textbf{78.86}
& \cellcolor{green!15}\textbf{57.78} \\

\hline
\end{tabular}%
\vspace{-1.0em}
\end{table*}

\noindent
\textbf{Implementation Details.} %
Unless otherwise specified, we pre-train MinkUNet-34~\cite{choy20194d}
on nuScenes using aligned image--LiDAR pairs, with a frozen
DINOv2~\cite{oquab2024dinov} ViT-L/14 or ViT-B/14 encoder as
the visual teacher, following~\cite{wozniak2026hilda}.
Pre-training is conducted for 25 epochs on four GPUs with a
batch size of 8 and a learning rate of $10^{-3}$.
We set the Poincaré ball curvature parameter to $c=0.1$ and
the distillation loss weights to $\lambda_{\mathrm{pp}}=1$,
$\lambda_{\mathrm{ss}}=1$, and $\lambda_{\mathrm{ctx}}=0.05$.
Superpixels are generated offline using SEEM~\cite{zou2023segment}
and reused throughout pre-training.

For downstream evaluation, linear probing and fine-tuning are conducted for 50 epochs on two GPUs. We use batch sizes of 32 for nuScenes and 12 for SemanticKITTI. The learning rate is set to $0.0005$, except for SemanticKITTI fine-tuning, where we use $0.001$. During downstream training, point clouds are augmented using random rotations about the $z$-axis and random flips along the $x$ and $y$ axes. Global scaling within $[0.9,1.1]$ is additionally applied for both linear probing and fine-tuning. Input images are resized to $224\times448$ without further image augmentation. For the domain adaptation experiments and selected ablations, we additionally use WaffleIron-768~\cite{Puy2023WI}, pretrained either on nuScenes or jointly across multiple datasets using randomly sampled aligned image--LiDAR pairs, following the backbone setting used for comparison with ScaLR.

For linear probing, the pre-trained backbone is kept frozen and only a batch normalization layer followed directly by a linear classification layer is optimized. For fine-tuning, the entire network is optimized end-to-end using varying proportions of labeled data. We compare \coolname{} against random initialization and eight state-of-the-art pre-training techniques, including PPKT, SLidR, Seal, Superflow, LargeAD, ScaLR, Superflow++ and LiMoE. Baseline results are taken directly from the corresponding official publications under matched evaluation settings. Performance is measured using mean Intersection-over-Union (mIoU) over all semantic classes for both linear probing and downstream semantic segmentation.

\noindent
\textbf{Benchmark Datasets.} We evaluate on nuScenes~\cite{caesar2020nuScenes}, SemanticKITTI (S.KITTI)~\cite{behley2019semantickitti}, and PandaSet considering its Pandar64 and PandarGT sensors separately~\cite{xiao2021pandaset}. %
We use the official train/validation splits and class definitions for nuScenes and SemanticKITTI. The datasets provide diverse LiDAR configurations, including the 32-beam Velodyne HDL-32E in nuScenes, 64-beam Velodyne in SemanticKITTI, 64-channel Pandar64, and a forward-facing PandarGT. We additionally evaluate corruption robustness on nuScenes-C from Robo3D~\cite{kong2023robo3d}, which contains eight corruption types representative of adverse driving conditions.

\section{Results} 
We comprehensively evaluate \coolname{} through linear probing, label-efficient fine-tuning, domain adaptation, qualitative analysis, and robustness to sensor corruptions. In addition, we study the impact of pretraining data scale and validate the hyperbolic confidence proxy.

\vspace{0.5em}
\textbf{Linear Probing and Label-Efficient Finetuning.}
We evaluate the quality and transferability of the learned LiDAR representations through linear probing, fine-tuning under varying annotation budgets, and cross-dataset transfer, with results summarized in Table~\ref{tab:finetuning_benchmark}. \coolname{} achieves 60.95\% and 60.07\% mIoU with ViT-B and ViT-L, respectively, under linear probing, surpassing the previous best \textit{ScaLR} by 9.05 and 8.30 percentage points. The confidence-aware hyperbolic image-to-LiDAR distillation provides reliable semantic supervision, yielding strong label efficiency across varying data ratios. In particular, \coolname{} achieves the best performance across all evaluated nuScenes data ratios, from 1\% to full supervision, for both ViT-B and ViT-L. Notably, \coolname{} requires only 25 pretraining epochs, compared with 50--150 epochs for competing methods, yet delivers substantial gains in the label-scarce regime. Under the 1\% annotation budget, \coolname{} records 52.46\% and 53.29\% mIoU with ViT-B and ViT-L, respectively, surpassing \textit{LiMoE} at 50.23\% and 51.41\%. For cross-dataset transfer, despite being pretrained solely on nuScenes, \coolname{} achieves 53.72\% and 57.78\% mIoU on SemanticKITTI with only 1\% labeled data using ViT-B and ViT-L, respectively, improving over the previous best \textit{LargeAD} by 3.04 and 6.10 points. These results demonstrate that \coolname{} learns transferable LiDAR representations that remain effective under both limited annotations and cross-dataset transfer.

\begin{table}[t!]
\centering
\caption{Comparative analysis of \textbf{Unsupervised Domain Adaptation} on four cross-sensor pairs, reporting mIoU (\%) for 10 shared classes. $\dagger$ indicates results from prior work. Modality (Mod.): U (Unimodal), M (Multimodal including Waymo), $\text{M}^*$ (Multimodal excluding Waymo). Datasets: \{NuScenes (N), SemanticKITTI (K), PandaGT, Panda64, Waymo (W)\}. Our approach achieves state-of-the-art average performance with a single pretrained backbone, without pretraining on Waymo.}
\label{tab:domain_adaptation}
\renewcommand{\arraystretch}{1.1}
\setlength{\tabcolsep}{3pt}
\small
\begin{tabular}{lc|cccc|c}
\hline
Method & Mod. & N$\rightarrow$K & K$\rightarrow$N & N$\rightarrow$W & W$\rightarrow$N & Avg. \\
\hline
Target oracle & - & 72.4 & 83.8 & 83.4 & 83.8 & 80.9 \\
Source only & - & 44.6 & 55.1 & 37.1 & 64.6 & 50.4 \\
\hline
PL$^\dagger$~\cite{morerio2017minimal} & U & 30.0 & 29.0 & 31.9 & 22.3 & 28.3 \\
CosMix$^\dagger$~\cite{saltori2022cosmix} & U & 30.6 & 29.7 & 31.5 & 30.0 & 30.5 \\
MM2D3D$^\dagger$~\cite{cardace2023exploiting} & M & 30.4 & 31.9 & 31.3 & 33.5 & 31.8 \\
MM2D3D$^{*\dagger}$~\cite{cardace2023exploiting} & M & 32.9 & 33.7 & 34.1 & 37.5 & 34.6 \\
Adapt-SAM$^\dagger$~\cite{peng2024learning} & M & \textbf{48.5} & 42.9 & 44.9 & 48.2 & 46.1 \\
MuDDoS$^\dagger$~\cite{Michele_2025_BMVC} & M & 41.8 & \underline{53.3} & \textbf{54.1} & \underline{53.0} & \underline{50.6} \\
\cellcolor{green!15}\textbf{\coolname{}}~(Ours) & \cellcolor{green!15} $\text{M}^*$ & \cellcolor{green!15} \underline{47.1} & \cellcolor{green!15} \textbf{59.8} & \cellcolor{green!15} \underline{49.6} & \cellcolor{green!15} \textbf{60.5} & \cellcolor{green!15} \textbf{54.2}\\
\hline
\end{tabular}
\vspace{-1.5em}
\end{table}

\begin{table*}[t]
    \centering
    \caption{\textbf{Robustness probing} of LiDAR pretraining methods under eight out-of-distribution corruptions in the \emph{nuScenes-C} dataset from the Robo3D benchmark \cite{kong2023robo3d}, using linear probing with frozen MinkUNet-34 backbones. \textbf{mCE}, \textbf{mRR}, and \textbf{mIoU} scores are reported in percentage (\%). \textbf{Avg} denotes the average mIoU across all eight corruptions. The \textbf{best} scores are highlighted in \textbf{bold}.}
\label{tab:robustness}
    \setlength{\tabcolsep}{6pt}
    \small
    \begin{tabular}{c|c|cc|p{0.8cm}<{\centering}p{0.8cm}<{\centering}p{0.8cm}<{\centering}p{0.8cm}<{\centering}p{0.8cm}<{\centering}p{0.8cm}<{\centering}p{0.8cm}<{\centering}p{0.8cm}<{\centering}|p{0.8cm}<{\centering}}
    \hline 
    \textbf{\#} & \textbf{Initial} & \textbf{mCE} $\downarrow$ & \textbf{mRR} $\uparrow$  & \textbf{Fog} & \textbf{Wet} & \textbf{Snow} & \textbf{Move} & \textbf{Beam} & \textbf{Cross} & \textbf{Echo} & \textbf{Sensor} & \textbf{Avg.}
    \\\hline
    \multirow{9}{*}{\rotatebox[origin=c]{90}{\textbf{LP}}}
    & PPKT~\cite{liu2021learning} & 183.44 & 78.15 & 30.65 & 35.42 & 28.12 & 29.21 & 32.82 & 19.52 & 28.01 & 20.71 & 28.06 \\
    & SLidR~\cite{sautier2022image} & 179.38 & 77.18 & 34.88 & 38.09 & 32.64 & 26.44 & 33.73 & 20.81 & 31.54 & 21.44 & 29.95 \\
    & Seal~\cite{liu2023segment} & 166.18 & 75.38 & 37.33 & 42.77 & 29.93 & 37.73 & 40.32 & 20.31 & 37.73 & 24.94 & 33.88 \\
    & Superflow~\cite{xu20244d} & 161.78 & 75.52 & 37.59 & 43.42 & 37.60 & 39.57 & 41.40 & 23.64 & 38.03 & 26.69 & 35.99 \\
    & ScaLR~\cite{puy2024three} & 173.18 & 78.91 & 37.55 & 37.96 & 36.29 & 33.64 & 33.06 & 23.01 & 33.62 & 23.70 & 33.59 \\
    & LargeAD~\cite{kong2025largead} & 160.28 & 77.29 & 40.54 & 43.26 & 37.92 & 38.27 & 40.27 & 25.67 & 39.26 & 30.62 & 36.98 \\
    & Superflow++~\cite{xu2025enhanced} & 158.98 & 75.70 & 38.52 & 44.36 & 37.91 & 41.32 & 41.68 & 24.25 & 40.22 & 27.65 & 36.99 \\
    & LiMoE~\cite{xu2025limoe} & 155.77 & 78.23 & 40.35 & 45.28 & 39.14 & 42.10 & 44.21 & 27.33 & 39.20 & 29.49 & 38.39 \\
    & \cellcolor{green!15}{\textbf{\coolname{}}~(Ours)} & \cellcolor{green!15}\textbf{126.69} & \cellcolor{green!15}\textbf{88.65} &  \cellcolor{green!15}\textbf{58.58} & \cellcolor{green!15}\textbf{58.98} & \cellcolor{green!15}\textbf{55.02} & \cellcolor{green!15}\textbf{51.69} & \cellcolor{green!15}\textbf{52.80} & \cellcolor{green!15}\textbf{57.55} & \cellcolor{green!15}\textbf{49.43} & \cellcolor{green!15}\textbf{41.48} & \cellcolor{green!15}\textbf{53.19} \\
    \hline
    \end{tabular}
\vspace{-1.0em}
\end{table*}

\vspace{0.5em}
\textbf{Domain Adaptation.} Following the MuDDoS evaluation protocol~\cite{Michele_2025_BMVC}, we benchmark unsupervised domain adaptation across four challenging cross-sensor scenarios (Table~\ref{tab:domain_adaptation}). \coolname{} achieves the highest average performance of 54.2\% mIoU, outperforming all prior methods, including the strongest baselines MuDDoS (50.6\%) and Adapt-SAM (46.1\%). The substantial improvements on K$\rightarrow$N (59.8\%) and W$\rightarrow$N (60.5\%) highlight the ability of \coolname{} to learn transferable representations that reduce domain discrepancy while preserving semantic structure. Notably, these gains are obtained using a single pre-trained backbone and without pre-training with Waymo~\cite{sun2020scalability}, underscoring the robustness and generalization capability of our adaptation framework.

\begin{figure}[t]
\centering
\includegraphics[width=.48\textwidth, scale=.9]{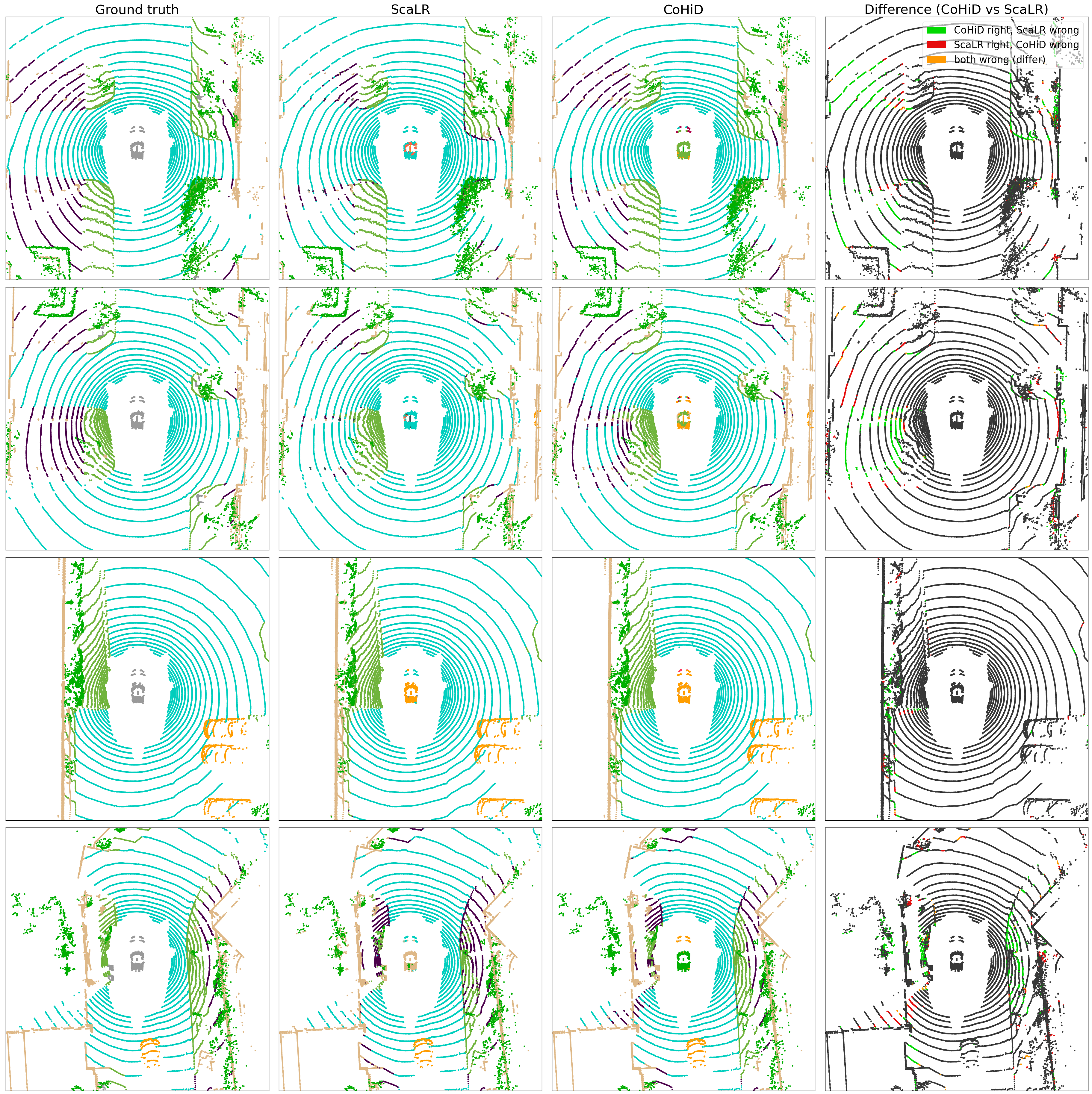}
\caption{Qualitative examples showing the improvements and failures of \coolname{} over ScaLR on the nuScenes validation set. Each row shows ground truth, ScaLR prediction, \coolname{} prediction, and disagreement map from left to right. Green, red, and orange points denote corrections, regressions, and shared failures, respectively. \coolname{} improves separation of similar classes and sparse long-range LiDAR predictions.}
\label{fig:nuscenes_best_top5}
\vspace{-1.5em}
\end{figure}

\vspace{0.5em}
\textbf{Qualitative Analysis.}
We qualitatively analyze cases where \coolname{} performs substantially better or worse than the ScaLR~\cite{puy2024three} baseline on the nuScenes validation set, shown in Fig. \ref{fig:nuscenes_best_top5}. In successful cases, \coolname{} resolves confusions between geometrically similar categories, such as vegetation versus terrain/manmade surfaces and sidewalk versus drivable areas, and improves predictions in sparse and long-range LiDAR regions where ScaLR tends to over-smooth. These gains indicate that hierarchy-aware contrastive distillation produces clearer semantic boundaries. In failure cases, \coolname{} struggles with weak or ambiguous geometric cues, particularly under occlusion, cluttered vegetation, and rare classes, leading to confusion among neighboring categories.

\vspace{0.5em}
\textbf{Robustness to Corruptions.} To evaluate robustness to out-of-distribution corruptions, we conduct experiments on the \emph{nuScenes-C} benchmark from Robo3D and report the results in Table~\ref{tab:robustness}. The benchmark includes eight challenging perturbations: fog, wet, snow, motion blur (Move), beam reduction, crosstalk, echo, and sensor failure. Under linear probing, \coolname{} achieves the best performance across all eight corruption types, attaining an average mIoU of 53.19\%, which is 14.80 points higher than the second-best method, LiMoE (38.39\%). \coolname{} further achieves an mRR of 88.65\% and an mCE of 126.69, demonstrating substantially stronger robustness under diverse environmental and sensor-induced corruptions. These consistent gains suggest that confidence-aware distillation provides more reliable cross-modal supervision, while hierarchical alignment promotes stable semantic representations beyond individual point-level correspondences, making the learned LiDAR features less sensitive to sensing degradations.

\begin{figure}[t]
\vspace{0.5em}
\centering
\begin{minipage}{\linewidth}
\centering
\begin{minipage}{0.49\linewidth}
    \centering
    \includegraphics[width=\linewidth]{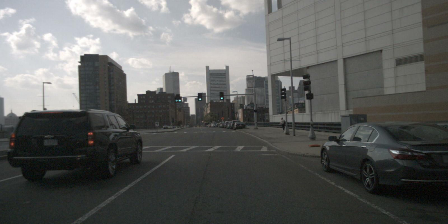}
\end{minipage}
\hfill
\begin{minipage}{0.49\linewidth}
    \centering
    \includegraphics[width=\linewidth]{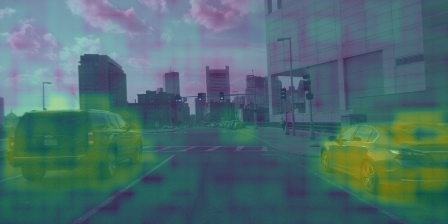}
\end{minipage}
\end{minipage}

\caption{Image-to-LiDAR alignment showing the input image (left image) and pixel-wise cosine similarity to a query LiDAR point from the car category (right image).
}
\label{fig:merged_3img}
\vspace{-1.5em}
\end{figure}

\vspace{0.5em}
\textbf{Effect of Pretraining Data Scale.} Table~\ref{tab:linprob_benchmark} reports linear probing results on semantic segmentation downstream tasks with the WI-768 backbone, where LiDAR features are distilled from a ViT-L/14 image encoder. When pretrained solely on nuScenes, our proposed method achieves state-of-the-art performance across all benchmarks, reaching 70.1\% on nuScenes, 47.4\% on SemanticKITTI, 35.8\% on Panda64, and 31.3\% on PandaGT, and achieving up to 4\% mIoU improvement over ScaLR. These gains highlight the effectiveness of confidence-aware distillation in filtering noisy pseudo-supervision and learning more robust 3D representations. Scaling pretraining to all available datasets further improves performance to 70.0\%, 57.3\%, 39.9\%, and 36.1\%, respectively, demonstrating that increased data diversity consistently enhances cross-dataset generalization. %
Complementing these quantitative results, Fig.~\ref{fig:merged_3img} illustrates the image-to-LiDAR alignment quality of our pretrained model. The left panel shows the input camera image, while the right panel visualizes the pixel-wise cosine similarity between a query LiDAR point on the left-side car and all image pixels. The resulting heatmap exhibits strong local coherence, with high similarity concentrated on both cars in the scene, indicating that features of the same object form coherent clusters in the joint embedding space.

\begin{table}[t!]
\centering
\caption{\textbf{Linear probing results} on the frozen, pretrained WI-768 backbone for each downstream dataset to assess representation quality. Performance is reported in mIoU (\%) on nuScenes, SemanticKITTI (S.KITTI), Panda64, and PandaGT. Our method consistently outperforms ScaLR in all settings.}
\label{tab:linprob_benchmark}
\resizebox{.49\textwidth}{!}{%
\centering
\small
\setlength{\tabcolsep}{2pt}
\begin{tabular}{cccccc}
\hline
\multirow{2}{*}{Method} & \multirow{2}{*}{Pretrain Dataset} & \multicolumn{4}{c}{Downstream \& Test Dataset} \\ \cdashline{3-6}
 &    & nuScenes &   S.KITTI  & Panda64 & PandaGT      \\ \hline
 ScaLR~\cite{puy2024three}    & nuscenes     &   67.8  &  43.1   &  33.9    &    29.9    \\

\cellcolor{green!15} \textbf{\coolname{}}~(Ours)        & \cellcolor{green!15}    nuScenes        & \cellcolor{green!15}  \textbf{70.1}  &  \cellcolor{green!15} \textbf{47.4}  &  \cellcolor{green!15}  \textbf{35.8}            &    \cellcolor{green!15}   \textbf{31.3}    \\

ScaLR~\cite{puy2024three}    & All Datasets     &   67.8  &  55.8   &  37.9    &    34.5    \\

\cellcolor{green!15} \textbf{\coolname{}}~(Ours)       & \cellcolor{green!15}    All Datasets        & \cellcolor{green!15}  \textbf{70.0}  &  \cellcolor{green!15} \textbf{57.3}  &  \cellcolor{green!15}  \textbf{39.9}            &    \cellcolor{green!15}   \textbf{36.1}    \\
 \hline

\end{tabular}%
}
\vspace{-1.5em}
\end{table}

\vspace{.5em}
\noindent

\begin{figure}[t]
    \centering
    \includegraphics[width=0.5\textwidth]{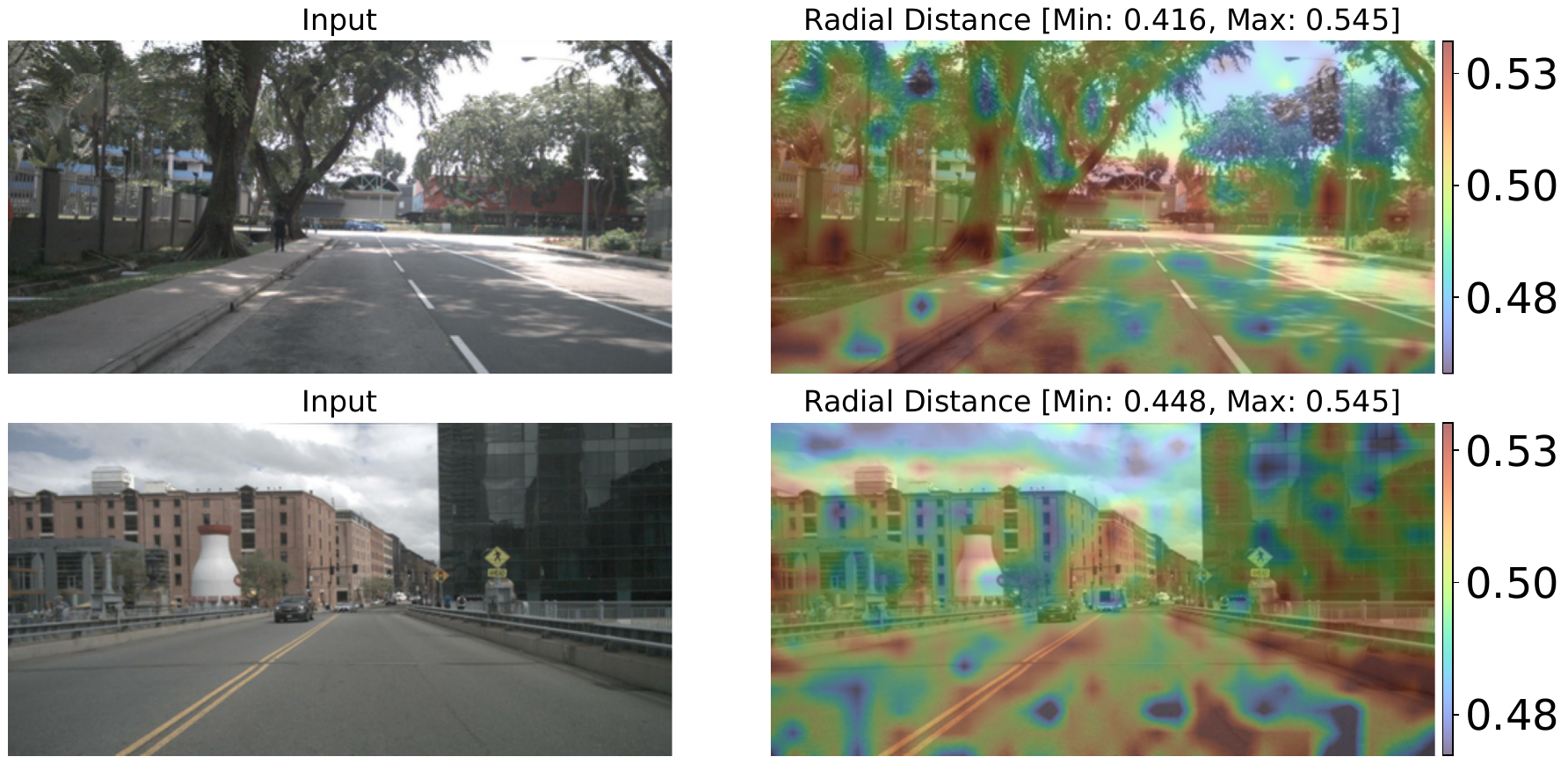} 
    \caption{Pixel-wise visualization of hyperbolic radial distance for two nuScenes camera frames. Left: input image. Right: overlay of $u(\mathbf{z}^I)=d_c(0,\mathbf{z}^I)$, obtained by mapping DINOv2 %
    features into the Poincaré ball and measuring their geodesic distance from the origin. Larger radii (warm colours) correspond to higher geometric confidence, whereas smaller radii (cool colours) correspond to lower confidence. Colour bar limits are the per-image minimum and maximum of $u$.}
    \label{fig:confidence_proxy}
    \vspace{-1.5em}
\end{figure}

\vspace{0.5em}
\textbf{Empirical validation of the hyperbolic confidence proxy.}
DINOv2 features can vary in reliability across image regions due to factors such as low texture, occlusion, and semantic ambiguity. Hence, assigning uniform weights to all image and LiDAR feature pairs during distillation may propagate unreliable image features to the LiDAR representation. To examine this variation, we visualise the pixel-wise hyperbolic radial distance $u(\mathbf{z}^I)=d_c(0,\mathbf{z}^I)$ of DINOv2 last-block features after mapping into the Poincaré ball $\mathbb{B}_c^d$ (Figure~\ref{fig:confidence_proxy}). The resulting maps show clear spatial variation, with structured regions often exhibiting larger radial distances (warm colours) and less distinctive regions exhibiting lower values (cool colours). Given that hyperbolic representational capacity increases with radial distance, this spatial variation motivates using $u(\mathbf{z}^I)$ as a geometry-derived confidence proxy. CoHiD therefore uses this signal to assign higher distillation weights to more reliable image andLiDAR feature pairs while reducing the influence of potentially noisy features, without requiring explicit confidence supervision.

\section{Ablation Study}

\vspace{-1.0em}
\begin{table}[b!]
\centering
\caption{\textbf{Component ablation of CoHiD on nuScenes.}
Each component is individually removed from the full model
using ViT-B/SEEM, while retaining the remaining components.
Results are reported in mIoU (\%) for linear probing (LP)
and $1\%$ fine-tuning (FT).
Conf. denotes hyperbolic confidence weighting.
Values in parentheses indicate performance changes relative
to the full model in percentage points.}
\label{tab:ablation_study}
\vspace{-0.5em}
\resizebox{\columnwidth}{!}{%
\begin{tabular}{cccccc}
\toprule
\multirow{2}{*}{$\mathcal{L}_{\mathrm{pp}}$} &
\multirow{2}{*}{$\mathcal{L}_{\mathrm{ss}}$} &
\multirow{2}{*}{$\mathcal{L}_{\mathrm{ctx}}$} &
\multirow{2}{*}{Conf.} &
\multicolumn{2}{c}{nuScenes (mIoU $\uparrow$)} \\
\cmidrule(lr){5-6}
& & & & LP & $1\%$ FT \\
\midrule
\xmark & \cmark & \cmark & \cmark
& 54.47 {\scriptsize($-6.48$)}
& 50.42 {\scriptsize($-2.04$)} \\
\cmark & \xmark & \cmark & \cmark
& 57.12 {\scriptsize($-3.83$)}
& 51.10 {\scriptsize($-1.36$)} \\
\cmark & \cmark & \xmark & \cmark
& 57.12 {\scriptsize($-3.83$)}
& 51.89 {\scriptsize($-0.57$)} \\
\cmark & \cmark & \cmark & \xmark
& 57.14 {\scriptsize($-3.81$)}
& 49.45 {\scriptsize($-3.01$)} \\
\midrule
\cmark & \cmark & \cmark & \cmark
& \textbf{60.95} & \textbf{52.46} \\
\bottomrule
\end{tabular}}
\vspace{-0.75em}
\end{table}

\vspace{1.5em}
\noindent\textbf{Component Ablation.}
Table~\ref{tab:ablation_study} reveals the distinct and
complementary roles of CoHiD's components in learning
transferable LiDAR representations.
Removing pixel-to-point distillation incurs the largest
degradation in linear probing ($6.48\%$), underscoring
the importance of fine-grained cross-modal correspondence
in preserving point-level semantic discriminability.
Removing either superpixel-to-superpoint or contextual
distillation reduces linear probing performance by
$3.83\%$, highlighting the importance of extending
knowledge transfer beyond local correspondences to
region-level semantics and global contextual relationships.
Interestingly, disabling hyperbolic confidence weighting
causes the largest degradation under $1\%$ fine-tuning
($3.01\%$), suggesting that selectively transferring
reliable semantic knowledge becomes particularly important
when downstream supervision is scarce.
These findings suggest that multi-level distillation
enriches the semantic structure of LiDAR representations,
while confidence-aware knowledge transfer enhances their
adaptability under limited supervision.

\section{Conclusion} 
\label{sec:conclusion}

This paper presents \coolname{}, a novel confidence-aware hierarchical distillation method for transferring semantic knowledge from images to LiDAR using hyperbolic space. By embedding cross-modal features in the Poincaré ball and weighting teacher supervision using radial confidence, \coolname{} selectively transfers informative visual features while preserving structured semantic relationships. %
Distillation is performed at pixel-to-point, superpixel-to-superpoint, and global scene levels, enabling complementary fine-grained, regional, and scene-level supervision without manual 3D annotations.
Experiments on semantic segmentation benchmarks demonstrate consistent improvements over prior methods, together with stronger robustness under corruption and domain shift. These results highlight the potential of hyperbolic geometry for scalable LiDAR representation learning in autonomous systems and robotics.

\balance{}
\bibliographystyle{IEEEtran}
\bibliography{references}

\end{document}